\documentclass[10pt,twocolumn,letterpaper]{article}

\usepackage[pagenumbers]{iccv} 

\newcommand{\EDN}{D-Net}
\newcommand{\VAN}{T-Net}

\newcommand{\argmax}{\mathop{\mathrm{argmax}}\limits}
\newcommand{\argmin}{\mathop{\mathrm{argmin}}\limits}

\definecolor{iccvblue}{rgb}{0.21,0.49,0.74}
\usepackage[pagebackref,breaklinks,colorlinks,allcolors=iccvblue]{hyperref}
\usepackage{pifont}
\newcommand{\cmark}{\ding{51}} 
\newcommand{\xmark}{\ding{55}} 

\def\paperID{1130} 
\def\confName{ICCV}
\def\confYear{2025}

\title{EvTurb: Event Camera Guided Turbulence Removal}

\author{
        {Yixing Liu$^{1,2\dagger}$~~~Minggui Teng$^{1,2\dagger}$~~~Yifei Xia$^{1,2}$~~~Peiqi Duan$^{1,2}$~~~Boxin Shi$^{1,2*}$}\\ {} \\
        {\small $^1$ State Key Laboratory of Multimedia Information Processing, School of Computer Science, Peking University}\\
        {\small $^2$ National Engineering Research Center of Visual Technology, School of Computer Science, Peking University}\\
        {\small \texttt{luiginixy@stu.pku.edu.cn} }
        {\small \texttt{\{minggui\_teng, yfxia, duanqi0001, shiboxin\}@pku.edu.cn} } \\
}

\begin{document}
\maketitle
\begin{abstract}
Atmospheric turbulence degrades image quality by introducing blur and geometric tilt distortions, posing significant challenges to downstream computer vision tasks. Existing single-image and multi-frame methods struggle with the highly ill-posed nature of this problem due to the compositional complexity of turbulence-induced distortions. To address this, we propose EvTurb, an event guided turbulence removal framework that leverages high-speed event streams to decouple blur and tilt effects. EvTurb decouples blur and tilt effects by modeling event-based turbulence formation, specifically through a novel two-step event-guided network: event integrals are first employed to reduce blur in the coarse outputs. This is followed by employing a variance map, derived from raw event streams, to eliminate the tilt distortion for the refined outputs. Additionally, we present TurbEvent, the first real-captured dataset featuring diverse turbulence scenarios. Experimental results demonstrate that EvTurb surpasses state-of-the-art methods while maintaining computational efficiency.
\end{abstract}
{\let\thefootnote\relax\footnotetext{$^\dagger$~Contributed equally to this work as first authors}
\let\thefootnote\relax\footnotetext{$^*$~Corresponding author}
\let\thefootnote\relax\footnotetext{$^*$~Project page: \url{https://github.com/ChipsAhoyM/EvTurb}}
}
\begin{figure}[t]
		\centering
		\includegraphics[width=\columnwidth]{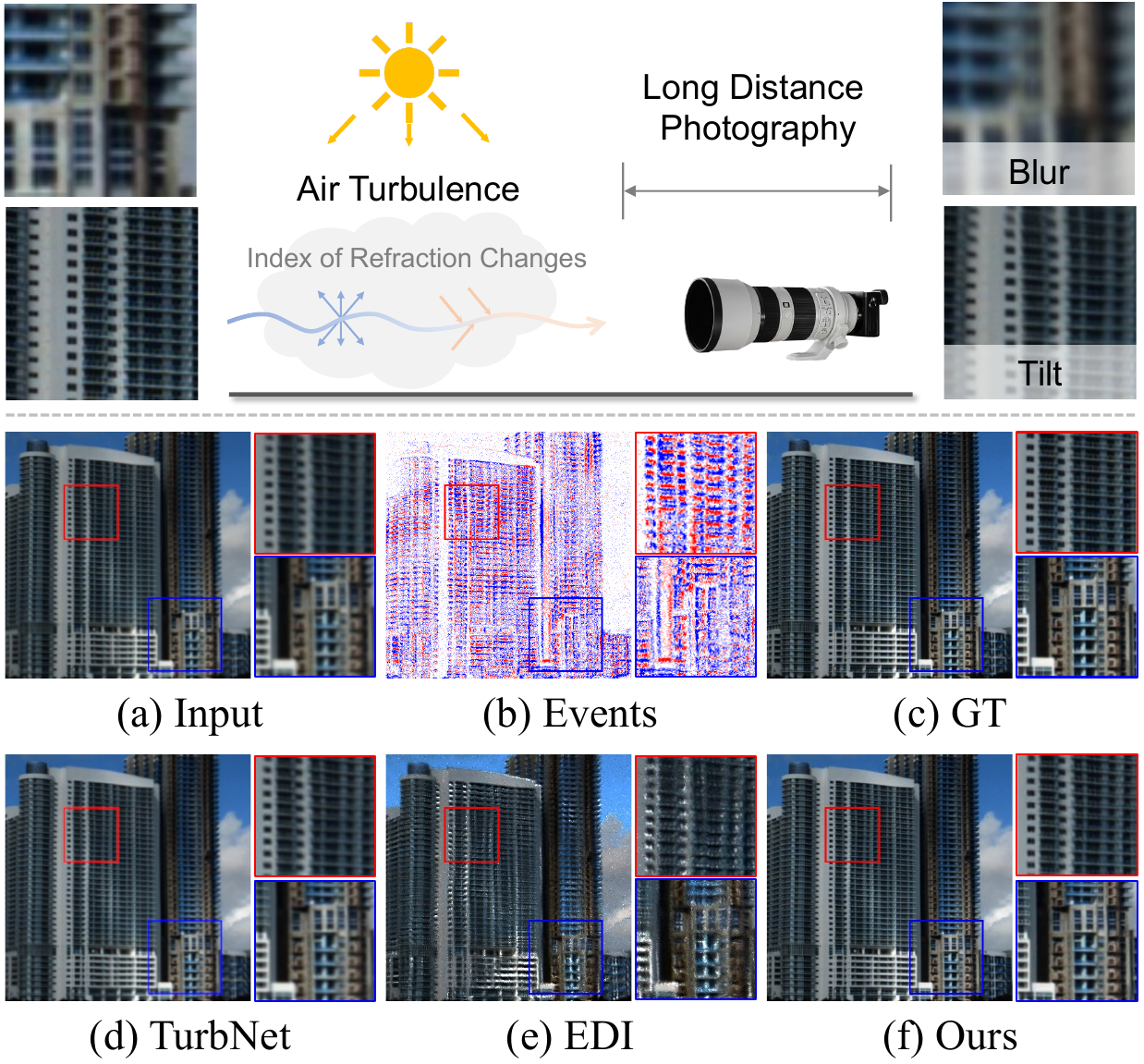}
        \caption{Turbulent images captured in long-distance photography often suffer from blur and geometric distortions due to atmospheric turbulence (top). Given a turbulent image (a) and corresponding events (b) recorded during exposure, our goal is to recover a clear image (c). The single-image method TurbNet~\cite{TurbNet} fails to fully correct distortions (d), while the event-based method EDI~\cite{pan2019bringing} introduces artifacts (e). In contrast, our EvTurb method produces a sharper image with fewer distortions (f).}
        \label{fig:teaser}
\end{figure}

\section{Introduction}\label{sec:intro}

Atmospheric turbulence refers to small-scale, irregular air motions, typically induced by heat, that degrade image quality through spatially varying changes in the index of refraction along the optical path. These variations introduce complex blur and geometric tilt effects (the top of \Cref{fig:teaser}), compounded over a fixed exposure duration. While recent methods~\cite{AT-Net,TurbNet} have been proposed for single-image atmospheric turbulence removal, their effectiveness remains limited~(\Cref{fig:teaser}(d)) due to the extremely ill-posed nature of the problem.

To address this challenge, multi-frame methods~\cite{Zhang_TMT,DATUM} leverage temporal information from image sequences, helping to mitigate the ill-posed nature of the problem (\eg, utilizing the lucky frame phenomenon~\cite{lucky}). While these approaches generally outperform single-image methods~\cite{AT-Net,TurbNet}, frame-based methods require large amounts of data and are further complicated by object motion. Computational photography methods~\cite{NB-GTR, steinbock2012adaptiveoptics} require additional hardware to achieve less turbulent outputs. For instance, narrow-band filters~\cite{NB-GTR} have been introduced to mitigate turbulence by integrating them with traditional RGB imaging. However, due to the constraints of traditional camera frame rates, tilt and blur effects are only partially reduced and remain compounded in the captured images.

Event cameras~\cite{serrano2013128event}, a type of neuromorphic sensor, generate continuous event streams when pixel radiance changes exceed preset thresholds. This capability enables them to detect variations with microsecond-level precision, supporting applications such as generating high-speed videos from event streams~\cite{tulyakov2021time,tulyakov2022time,rebecq2019high,teng2022nest}. This feature further facilitates high-speed observation of dynamic scenes, which motivates us to explore: \textit{Can event cameras capture turbulence field variations and effectively model blur and tilt effects separately using high-speed event signals?}

\begin{figure}
    \centering
    \includegraphics[width=\linewidth]{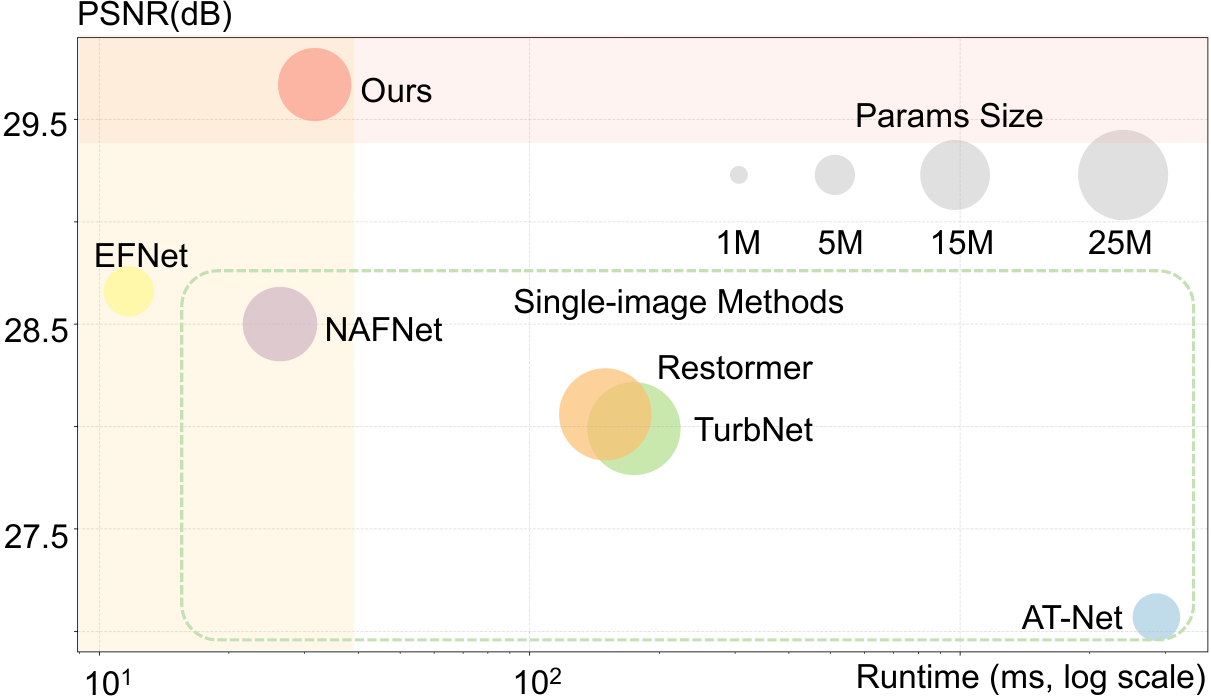}
    \caption{Comparison of PSNR, runtime, and model size. PSNR is evaluated on the TurbEvent dataset, and runtime is measured on an NVIDIA RTX 3090 GPU. By modeling the physics of event-based turbulence and leveraging high-speed information, our method achieves the highest PSNR with competitive runtime and compact model size.}
    \label{fig:runtime}
\end{figure}

In this paper, we propose the \textbf{EvTurb} framework, an \textbf{Ev}ent camera-guided \textbf{Turb}ulence removal approach designed to mitigate atmospheric turbulence effects in a single image by leveraging high-speed event streams. Event cameras capture variations in the turbulence field, recording both the intensity and frequency of these variations within the exposure time. However, existing event-based deblurring models account only for blur effects and cannot be directly applied to this scenario with coupled tilt effects, further introducing artifacts (\Cref{fig:teaser}(e)). To address this issue, we formulate an event-based image turbulence formation model, where the blur operator is derived from event-recorded intensity changes, while the tilt operator is characterized by event-triggering frequency. Events provide a high-fidelity representation of turbulence field variance, allowing for the separate processing of blur and tilt effects, thereby reducing the ill-posedness of the problem. Building on this model, we propose a two-step event-guided turbulence removal network that utilizes event temporal information. First, event integrals, representing the intensity of turbulence field variations, are fused with RGB images to generate coarse results that primarily mitigate blur. Next, a variance map computed from raw event streams captures the frequency of these variations and is integrated with the coarse results to remove tilt distortions, producing a refined final output (\Cref{fig:teaser}(f)). Overall, our contributions include:
\begin{itemize}
\item establishing a physical connection between high-speed event data and turbulence-distorted images by decoupling blur and tilt effects;
\item proposing a two-step framework that integrates turbulence field variations from event data and RGB images to separately address blur and tilt distortions; and
\item curating the first real-captured TurbEvent benchmarking dataset, featuring diverse turbulence patterns for comprehensive evaluation.
\end{itemize}
Quantitative and qualitative results on our newly collected TurbEvent dataset demonstrate that our method surpasses state-of-the-art techniques in recovering turbulence-free images with reduced distortion. As shown in \Cref{fig:runtime}, our method achieves an improvement of over $1$~dB in PSNR while maintaining a relatively short runtime.
\section{Related Work} \label{sec:related}
\paragraph{Atmospheric turbulence mitigation.}
The study of turbulence mitigation has long relied on traditional methods such as deconvolution for images with a limited field of view, assuming spatially invariant turbulence~\cite{Deconv}, and selecting lucky frames when addressing spatially variant turbulence~\cite{lucky, Vorontsov:01}. More recently, deep learning approaches have emerged, categorized into single-frame and multi-frame methods. Single-frame strategies employ backbones including CNNs~\cite{CNN-face}, Transformers~\cite{TurbNet}, Generative Adversarial Networks (GANs)~\cite{LTT-GAN, AT-FaceGAN, DT-GAN, TSRWGAN}, and Denoising Diffusion Probabilistic Models for turbulence correction~\cite{AT-DDPM}, alongside methods integrating RGB with narrow-band images~\cite{NB-GTR}. However, these techniques often overlook crucial temporal information. In contrast, multi-frame methods effectively utilize temporal data, either using transformer based architectures~\cite{Zhang_TMT, LRTM} or recurrent models~\cite{DATUM}. Some methods use special tricks to facilitate this process, such as statistical prior~\cite{LRTM}, neural representation~\cite{Nert, AT-NER} or in specail domain (\eg, planetary images~\cite{xia2025planet}).  Despite their efficacy, they require substantial datasets of $10$ to $12$ turbulent frames and face challenges related to varying turbulence and object motion between frames.

\paragraph{Event-based deblurring.}
Event-based image deblurring exploits the intrinsic connection between motion events and blurry images to reconstruct sharp details. A basic model of this field is the Event-based Double Integral (EDI) model~\cite{pan2019bringing}, which represents event streams and sharp/blurry images in the log domain. To improve robustness against noise and enhance fine details, CNN-based methods were later introduced~\cite{chen2023learning, zhou2023deblurringlowlight}. More recently, transformer architectures~\cite{sun2022efnet,sun2023event} have been leveraged, allowing attention mechanisms to better bridge the domain gap between event data and images, leading to more precise reconstructions. Additionally, event representations~\cite{teng2022nest} were explored to achieve a better fusion of image and event modalities. To address the challenge of image and event data domain gap, cross-modal calibration strategy was incorporated to address spatial and temporal alignment~\cite{yang2024motion}.

Unlike conventional cameras, event cameras directly record brightness changes, making them well-suited for capturing turbulence variations. While \citet{boehrer2021turbulence} use events to reconstruct high-speed videos and apply video-based turbulence removal methods, they do not explicitly model the relationship between events and turbulence images. Thus, we integrate motion cues from event streams into turbulence removal frameworks at the pixel level to more effectively leverage the rich temporal information in continuous event data for turbulence removal.

\section{Method}\label{sec:method}
In this section, we first introduce the concepts of event cameras and event-based turbulence formation in \Cref{sec:pre}. Next, we present a two-step strategy for reconstructing turbulence-free images, guided by turbulence information encoded in event streams, in \Cref{sec:evturb}. We then introduce our EvTurb framework in \Cref{sec:network}. The curation process of the TurbEvent dataset is described in \Cref{sec:dataset}, followed by implementation details in \Cref{sec:implement}.

\subsection{Formulations}\label{sec:pre}
\paragraph{Clear image formation with events.}
Event cameras~\cite{serrano2013128event} are designed to detect changes in radiance in the logarithmic domain. When the intensity change surpasses a preset threshold $c$, an event signal is generated as a quadruple, denoted as $(x, y, t, p)$, \ie,
\begin{align}\label{eq:ev}
   \log(\mathbf{I}^{t}_{x,y}) - \log(\mathbf{I}^{t_0}_{x,y})  = p \cdot c,
\end{align}
where $\mathbf{I}^{t}_{x,y}$ and $\mathbf{I}^{t_0}_{x,y}$ denote the intensity values of a pixel of $(x, y)$ at times $t$ and $t_0$, respectively. The polarity $p \in \{1, -1\}$ indicates whether the intensity is increasing or decreasing. Since \Cref{eq:ev} is applied independently for each pixel, the pixel indices are omitted henceforth.

Given two consecutive clear images $\mathbf{I}^{0}$ and $\mathbf{I}^{1}$ without turbulence, along with the corresponding $N_e$ events $\{e_k\}^{N_e}$ triggered between them, we can connect these two frames using the event integral:
\begin{align}\label{eq:evframe}
     \mathbf{I}^{1} = \mathbf{I}^{0} \cdot \exp(\int_{t_0}^{t_1} c_k \cdot p_k~\text{d}k),
\end{align}
where $c_k$ represents the spatial-temporal variant threshold, which is dependent on the scene conditions~\cite{hu2021v2e}.

\begin{figure*}
		\centering
		\includegraphics[width=\linewidth]{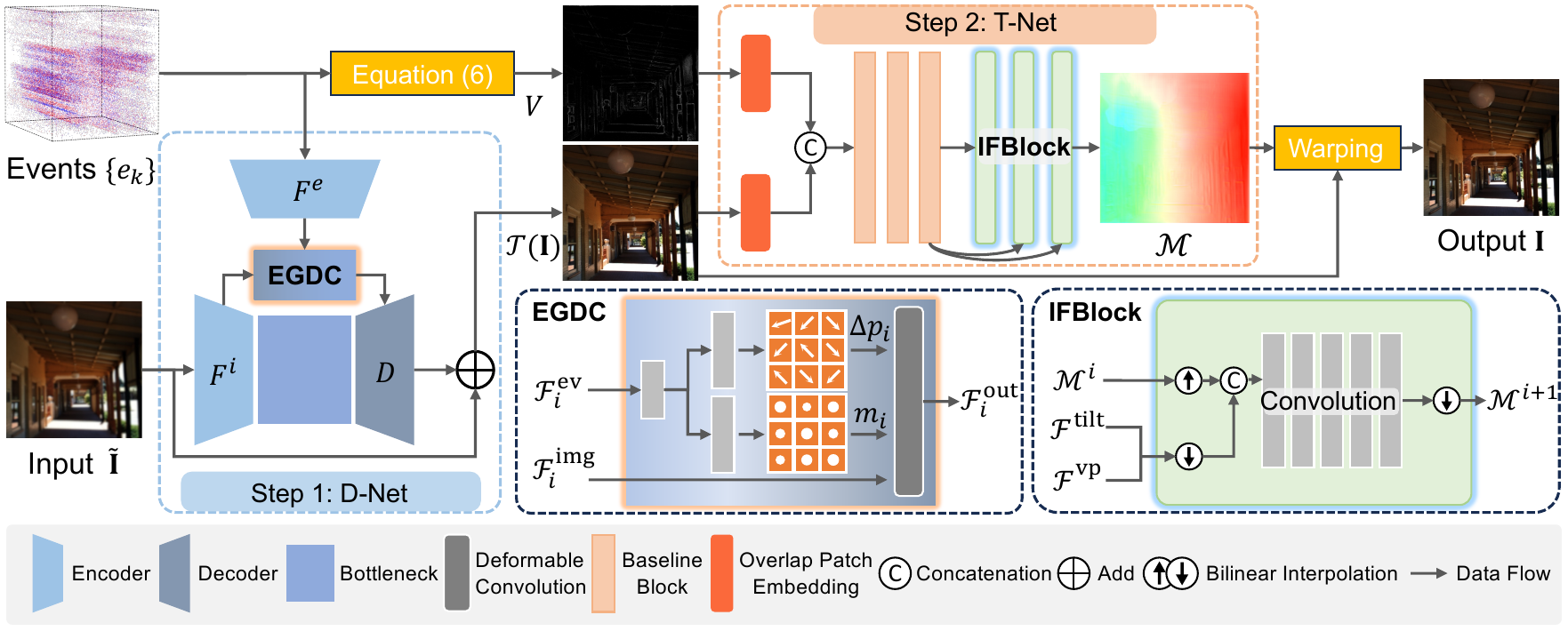}
		\caption{The overall framework of our EvTurb method. We first perform deblurring using a multi-scale architecture, D-Net, which estimates the event integral from two modalities: the RGB image $\tilde{\mathbf{I}}$ and the event stream $\{e_k\}$. The event-guided deformable convolution (EGDC) block is applied during the decoding process to enhance data fusion. After obtaining the blur-free image $\mathcal{T}({\mathbf{I}})$, T-Net predicts the tilt flow $\mathcal{M}$ guided by the calculated variance map $\mathbf{V}$. The IFBlock is then used to iteratively refine the flow at lower resolutions. Finally, the estimated flow $\mathcal{M}$ is used to warp $\mathcal{T}({\mathbf{I}})$, generating the output $\mathbf{I}$.}
        \label{fig:pipeline}
\end{figure*}

\paragraph{Turbulent image formation with events.} 
A turbulent image, denoted as $\tilde{\mathbf{I}}$, can be generated using the classical split-step wave-propagation equation. For simplicity, this process can be implemented as a forward equation~\cite{TurbNet}, approximating turbulence distortion as a combination of two degradation operators:
\begin{align}\label{eq:turb}
    \tilde{\mathbf{I}} = \mathcal{B} \circ \mathcal{T} (\mathbf{I}) + \mathbf{N},
\end{align}
where $\mathcal{B}$ represents the spatially varying blur, $\mathcal{T}$ denotes the geometric pixel displacement (commonly known as the tilt). An example is shown in \Cref{fig:teaser}. $\mathbf{N}$ is the additive noise signal, and $\circ$ denotes functional composition. If only a single turbulent image is available, the blur and tilt operations are inseparable, as these distortions are coupled together during the exposure time.

As demonstrated in \Cref{eq:evframe}, the latent images can be effectively linked with event streams. In scenarios of atmospheric turbulence, since tilt distortion predominantly arises due to air refraction, the high-speed latent images are less blurred but tilt distorted images. Furthermore, the blur effect can predominantly be considered as an average across latent frames. By integrating \Cref{eq:evframe} with \Cref{eq:turb}, we derive the following equation:
\begin{align}\label{eq:evdeturb} 
\tilde{\mathbf{I}} &\approx \mathcal{T} (\mathbf{I}) \cdot  \frac{1}{T}\int_T \left( \exp( \int_t c_k\cdot p_k~\text{d}k) \right )\text{d}t + \mathbf{N}\notag \\
& = \mathcal{T} (\mathbf{I}) \cdot \mathbf{E} + \mathbf{N}.
\end{align}
The blur operator is substituted with $\mathbf{E}$, which represents a double integral over the events occurring within the exposure period $T$. Note that our blur operator is distinct from standard motion blur. Leveraging the ability of high-speed event cameras to capture defocus kernels~\cite{lou2023efs}, our blur operator integrates local defocus effects, which can be further refined using event-based guidance.

\subsection{Event guided turbulence removal}\label{sec:evturb}
\paragraph{Blur removal.}  The formation model \Cref{eq:evdeturb} shows that turbulence distortion is compositional, comprising tilt and blur. By leveraging the high-speed turbulence field information captured in the event stream, the blur operator can be formulated as a double integral of events and subsequently canceled out from turbulence-distorted images. Recognizing the presence of noise, we further transform the \Cref{eq:evdeturb} into a least squares optimization problem:
\begin{align} \label{eq:argmin}
\mathcal{T} (\mathbf{I})^*  = \argmin_{\mathcal{T} (\mathbf{I})} \Vert \tilde{\mathbf{I}} - \mathcal{T} (\mathbf{I}) \cdot \mathbf{E} \Vert^2. 
\end{align}
Solving this objective facilitates the initial recovery of images $\mathcal{T} (\mathbf{I})^*$ that are less blurred yet exhibit tilt distortion.

\paragraph{Tilt removal.} Since the event integral is also affected by tilt distortion, it is necessary to explore the temporal information within the event stream to further enhance tilt removal. Inspired by previous work~\cite{AT-Net}, we recognize that variance maps can indicate pixel uncertainty related to tilt intensity. We construct a variance map from the raw event streams,  capturing the frequency of variations in the turbulence field and indicating the tilt distortion uncertainty. As events record the intensity value changes of latent frames, we define the variance of the entire latent frame sequence as the variance map. Formally, we calculate the variance map as follows:
\begin{equation}\label{eq:vp}
    \mathbf{V} = \texttt{norm}\left(\frac{1}{N_e} \sum_{i=1}^{N_e} \mathcal{E}_i^2 -  (\frac{1}{N_e} \sum_{i=1}^{N_e} \mathcal{E}_i)^2 \right),
\end{equation}
where \texttt{norm} denotes the normalization operation and $\mathcal{E}_i  = \sum_{k=1}^{i} \exp(c_k \cdot e_k)$ represents the accumulated events. 

Utilizing the variance map $\mathbf{V}$ as guidance, tilt flow $\mathcal{M}$ can be formulated under Maximum-a-Posteriori:
\begin{align}\label{eq:argmax}
    \mathcal{M}^* = \argmax_{\mathcal{M}}\mathbb{P}(\mathcal{M} \circ \mathcal{T} (\mathbf{I})^* \mid \mathbf{V}),
\end{align}
where $\mathbb{P}(\cdot)$ represents the distribution of natural images without tilt effects, and $\circ$ denotes the image warping function. By estimating the optimal tilt flow $\mathcal{M}^*$ using natural image priors, we can effectively apply it to the initial image for the tilt correction.

\subsection{EvTurb Framework}\label{sec:network}
\paragraph{\EDN.} The key component for optimizing \Cref{eq:argmin} is to obtain an accurate event integral. Since the thresholds of event cameras exhibit spatial-temporal variations~\cite{hu2021v2e}, directly integrating raw event data introduces artifacts into the restored images. Instead, we propose estimating this integral in a data-driven manner, converting \Cref{eq:argmin} into a regression problem, \ie:
\begin{equation}
    \mathcal{T}(\mathbf{I}) = D (F^i(\tilde{\mathbf{I}}), F^e(\{e_k\})),
\end{equation}
where $F^i$ and $F^e$ are implicit functions for encoding images and events, respectively, and $D$ is an implicit function for decoding and fusing information from both modalities.

To enhance feature extraction from both event data and RGB images, we propose a dual-branch network, \EDN, specifically designed to fuse these two modalities effectively. Prior studies~\cite{chen2023learning,zhou2023deblurringlowlight} have validated that the U-Net architecture is highly effective in restoring sharp image details due to its capability of extracting features at various scales. Therefore, we adopt a multi-scale U-Net structure for fusing the two modalities at different scales, and we employ dense blocks within both encoders to effectively reuse refined features.

To further improve local feature restoration, we introduce an event-guided deformable convolution (EGDC) module, incorporating deformable convolution into the skip connections. The polarity of event data serves as an important indicator for offset estimation, while accumulated intensity highlights significant areas, effectively functioning as a mask. Specifically, event features at certain scales are utilized to compute the sampling point offsets and the mask map as follows:
\begin{equation}
    \Delta p_{i} = C_{p}(B(\mathcal{F}_{i}^{\texttt{ev}})), \quad m_i = C_{m}(B(\mathcal{F}_{i}^{\texttt{ev}})),
\end{equation}
where $\Delta p_{i}$, $m_i$, and $\mathcal{F}_{i}^{\texttt{ev}}$ represent the offset map, mask map, and the $i$-th event feature, respectively. Here, $B$ denotes a bottleneck convolution, while $C_{p}$ and $C_{m}$ represent two separate convolutional layers. Using these computed offsets and masks, the deformable convolution can be calculated as:
\begin{equation}
    \mathcal{F}_{i}^{\texttt{out}} = \texttt{Deform}(\mathcal{F}_i^{\texttt{img}}, \Delta p_{i}, m_i).
\end{equation}
where \texttt{Deform} represents deformable convolution~\cite{DeformConv}. By employing this event-guided deformable convolution, we effectively address the issue of salient local features and align gradients within the feature space.

After estimating the event integral, we fuse it with turbulent images through a global connection, effectively generating deblurred outputs. Specifically, the global connection allows the model to leverage long-range dependencies between event estimation and turbulent image features, thereby effectively mitigating blur.

\paragraph{\VAN.} 
Despite obtaining coarse results from \EDN, the tilt effect still remained. Therefore, an essential step in \Cref{eq:argmin} for reconstructing the image is fitting a tilt-free natural image prior. Drawing inspiration from previous works on solving spatially-variant shifts~\cite{IF-Net, EvUnroll}, we propose a flow-based network, \VAN, to estimate the tilt flow $\mathcal{M}$, which warps the distorted image into a tilt-free image. And \Cref{eq:argmax} can be formulated as:
\begin{equation}
    \mathcal{M} = f_{\text{VT}}(\mathcal{T}(\mathbf{I}), \mathbf{V}),
\end{equation}
where $f_{\text{VT}}$ is an implicit function represented by \VAN, specifically designed for tilt flow estimation, and $\mathbf{V}$ is the calculated variance map derived from \Cref{eq:vp}, representing the uncertainty introduced by turbulence.

In designing~\VAN, we utilize overlapped patch embedding, which is adept at extracting contextual and boundary information. This approach ensures smooth transitions between patches and reducing artifacts in the final output. Subsequently, we employ a sequence of convolutions and Baseline Blocks~\cite{NAFNet} to effectively fuse features from two modalities. The next step is to determine a tilt flow from the fused latent features that accurately warps the distorted image toward the ground truth. To achieve this, we adopt IFBlocks~\cite{IF-Net} for flow estimation, iteratively refining and approaching the correct tilt flow. Formally, this iterative refinement can be expressed as:
\begin{equation}
    \mathcal{M}^{i+1} = \texttt{IF}(\mathcal{F}^{\texttt{tilt}}, \mathcal{F}^{\texttt{vp}}, \mathcal{M}^i),
\end{equation}
where $\mathcal{F}^{\texttt{tilt}}$ and $\mathcal{F}^{\texttt{vp}}$ denote extracted features from the deblurred coarse image and variance map, respectively, and \texttt{IF} represents the IFBlock. As shown in \Cref{fig:pipeline}, we computed tilt flow in the lower resolution level to ensure the smoothness of the tilt flow. By applying the final estimated flow to the coarse results obtained from~\EDN, we can reconstruct turbulence-free images.

\begin{figure}
    \centering
    \includegraphics[width=\linewidth]{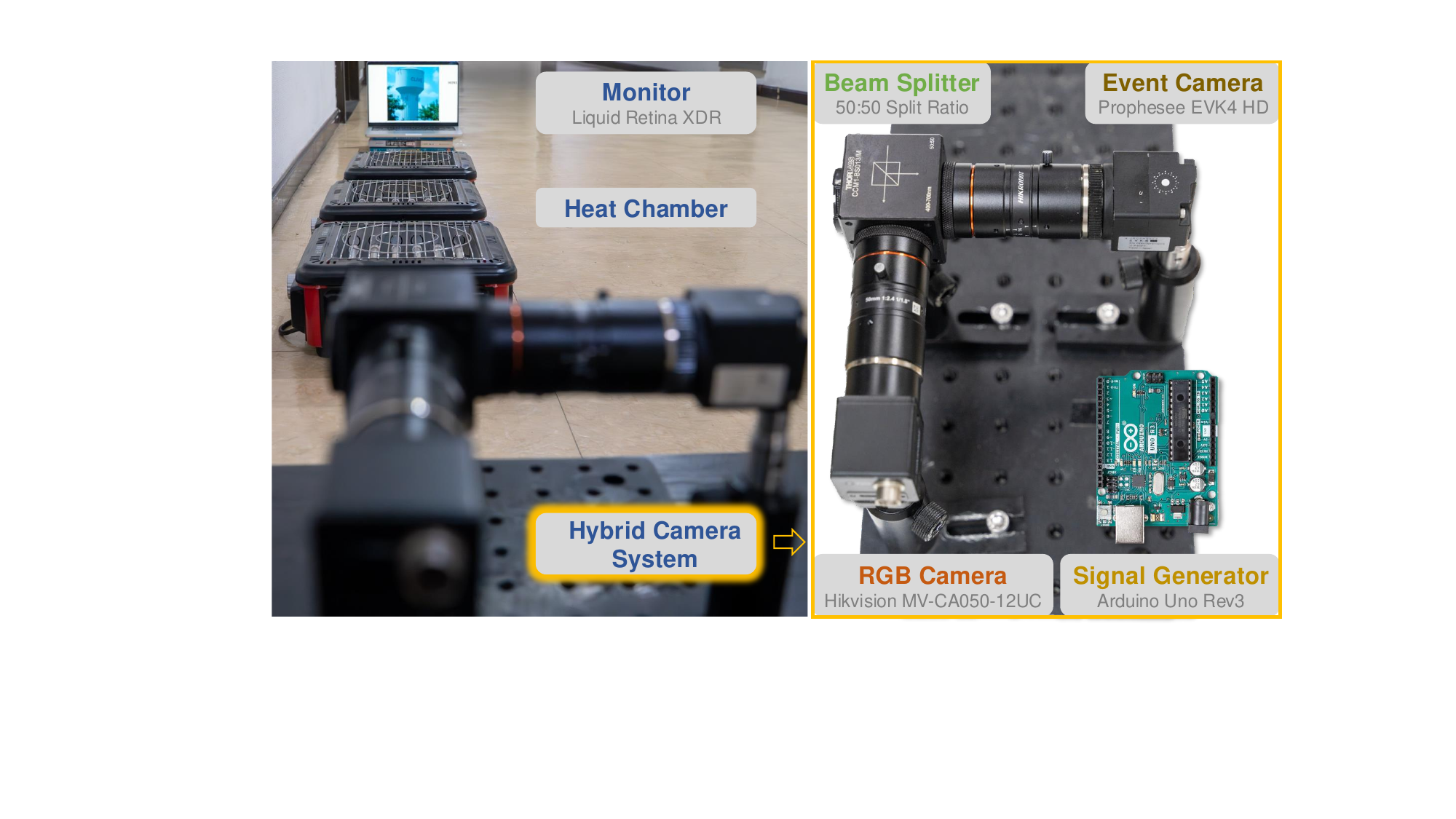}
    \caption{We construct a hybrid camera system (a) to investigate the relationship between event formation and image turbulence. Additionally, we implement a display-camera system with a heat chamber (b) for the curation of the TurbEvent dataset.}
    \label{fig:system}
\end{figure}

\subsection{TurbEvent dataset}\label{sec:dataset}
Simulating turbulence as a continuous process is challenging due to the lack of a closed-form formulation. This difficulty extends to generating high-frame-rate videos for simulated event streams. Inspired by the EventNFS dataset~\cite{duan2021eventzoom}, we designed a display-camera system to capture turbulent images, paired turbulence-free images, and corresponding event streams. 


Our setup includes a Liquid Retina XDR display (resolution: $3024\times 1964$, $120$Hz) and a hybrid camera system, shown in \Cref{fig:system}. The hybrid system consists of two cameras: a machine vision camera (HIKVISION MV-CA050-12UC) and an event camera (PROPHESEE GEN4.0), both equipped with $50$mm F/$1.8$ lenses and aligned using a beam splitter. To generate air turbulence, we employ multiple heat chambers, positioning the hybrid camera system approximately five meters from the display. Spatial calibration is performed using a checkerboard, while temporal synchronization is achieved via a signal generator.

We select images from the ADE20K dataset~\cite{zhou2017scene, zhou2019semantic}. This dataset spans multiple categories, including both outdoor and indoor scenes, and provides fine-grained details for robust analysis. The videos are played at $10$ FPS to avoid frame drops and minimize the effects of display refreshing. To ensure data integrity, influence from external light sources is minimized during recording. Ultimately, we successfully obtain $6318$ pairs of turbulent and turbulence-free images along with corresponding event streams, featuring $592\times592$ resolution. We refer to this event-based turbulence dataset as the ``TurbEvent" dataset. \footnote{Dataset configurations can be found in the supplementary.}

\begin{figure}
    \centering
    \includegraphics[width=\linewidth]{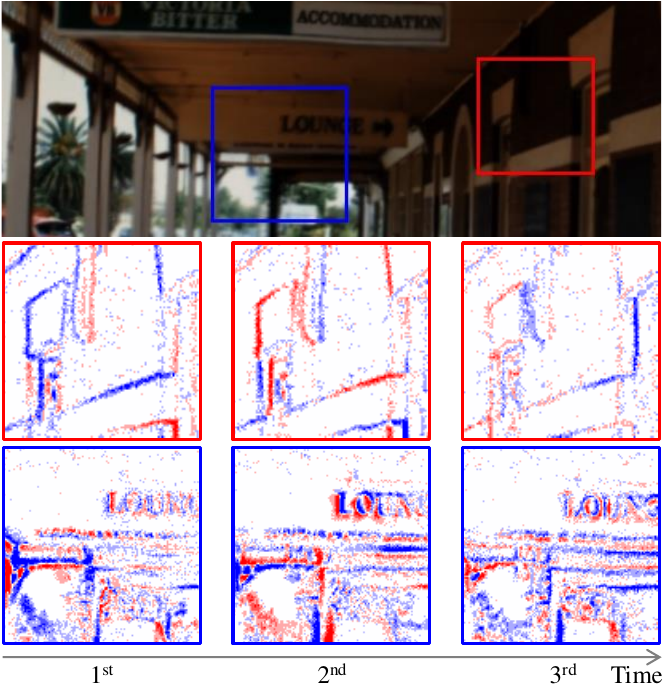}
    \caption{Visualization of our TurbEvent turbulent video frame (upper) and three repeated event recordings (lower) of the turbulent video. The turbulence patterns in these three event clips are distinct and exhibit randomness.}
    \label{fig:turpattern}
\end{figure}

\paragraph{Turbulence pattern.}
An example is presented in \Cref{fig:turpattern}, depicting a frame from the ADE20K dataset~\cite{zhou2017scene, zhou2019semantic} alongside its corresponding event patches, which were captured at continuous timestamps using our hybrid camera system. We consecutively record three event streams under the same scenario. Despite the fixed capturing conditions, the three event patches exhibit varying appearances due to the randomness of turbulence. Some edge signals are missing, while other events exhibit polarity reversal at the same position across three captures due to the varying and random nature of the turbulence pattern. By capturing a substantial amount of data, we are able to collect diverse turbulence patterns in the TurbEvent dataset. This inherent randomness makes it challenging to generate realistic simulated events from high-speed turbulent video simulations, reinforcing our decision to capture a real-world dataset.

\begin{table}[t]
		\centering
		\small
        \caption{Quantitative comparisons on TurbEvent dataset. The ``Event" column specifies if methods require events (Yes [\cmark] or No [\xmark]). $\uparrow$ ($\downarrow$) indicates the higher (lower), the better.}
\label{tab:heat}
\begin{tabular}{lcccc}
    \toprule
    & Event & ~PSNR $\uparrow$  & ~SSIM $\uparrow$ & ~LPIPS $\downarrow$ \\
    \midrule
    ATNet & \xmark & 27.07 & 0.7491 & 0.3416 \\
    TurbNet & \xmark & 27.99 & 0.7872 & 0.3418 \\
    N-DIR & \xmark & 23.19 & 0.6941 & 0.4560 \\
    NAFNet & \xmark & 28.50 & 0.8190 & 0.3218 \\
    Restormer & \xmark & 28.06 & 0.7995 & 0.2163 \\
    \midrule
    Restormer$^*$ & \cmark & 28.63 & 0.8114 & 0.2122 \\
    EFNet & \cmark & 28.66 & 0.8095 & 0.2140 \\
    Ours & \cmark & 29.67 & 0.8405 & 0.2010 \\
    \bottomrule
    \end{tabular}
    
\end{table}

\begin{figure*}[t]
		\centering
		\includegraphics[width=1.0\linewidth]{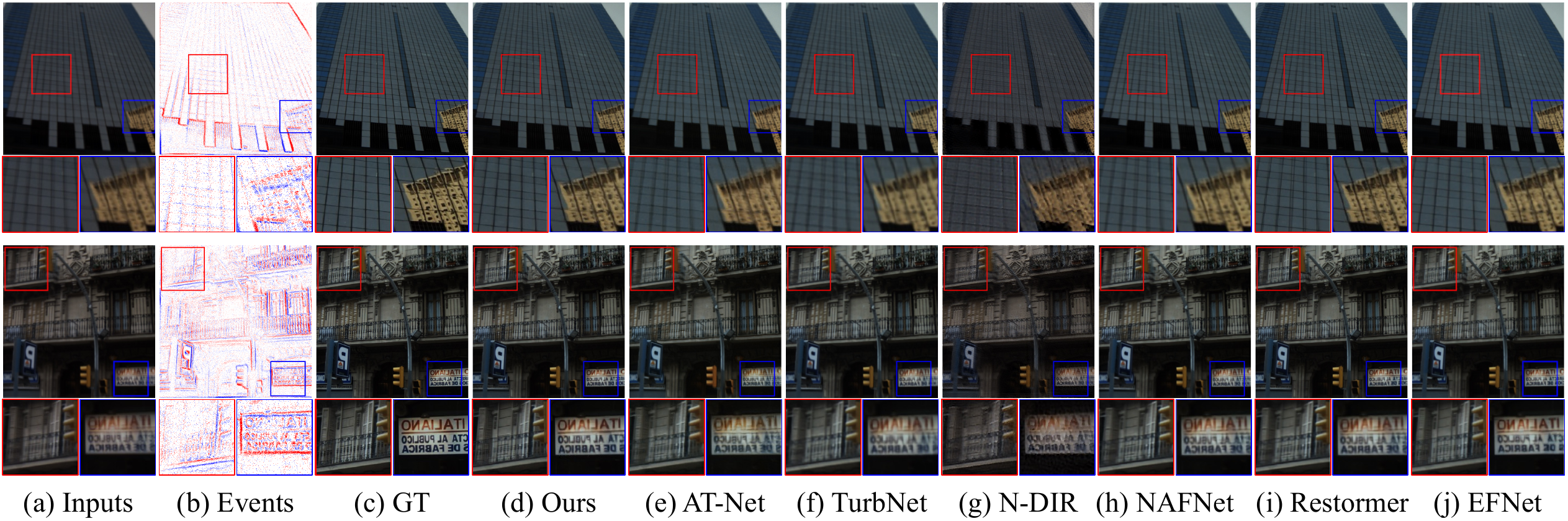}
		\caption{Visual quality comparison on the TurbEvent dataset. (a) Input images. (b) Events. (c) Ground truth. (d)$\sim$(j) Results of ours, AT-Net~\cite{AT-Net}, TurbNet~\cite{TurbNet}, N-DIR~\cite{Li_2021_NIDR}, NAFNet~\cite{NAFNet}, Restormer~\cite{Zamir2021Restormer}, and EFNet~\cite{sun2022efnet}. Close-up views are provided below each image. Additional results are provided in the supplementary material.} 
		\label{fig:realdata}
\end{figure*}

\begin{figure*}[t]
		\centering
		\includegraphics[width=1.0\linewidth]{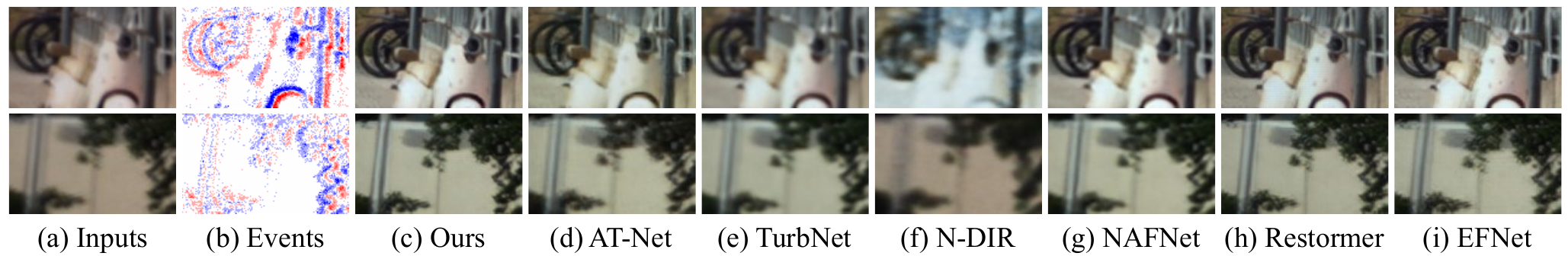}
		\caption{Visual quality comparison on real-captured data. (a) Input images. (b) Events. (c)$\sim$(i) Results of ours, AT-Net~\cite{AT-Net}, TurbNet~\cite{TurbNet}, N-DIR~\cite{Li_2021_NIDR}, NAFNet~\cite{NAFNet}, Restormer~\cite{Zamir2021Restormer}, and EFNet~\cite{sun2022efnet}. Additional results are provided in the supplementary material.} 
		\label{fig:realdata2}
\end{figure*}

\subsection{Implementation details}\label{sec:implement}
\paragraph{Loss function.}
We utilize a weighted combination of Mean Square Error (MSE) loss and perceptual loss~\cite{Perceptual} for D-Net and T-Net training:
\begin{equation}
\mathcal{L} = \lambda_1 \mathcal{L}_2(I_G, I_O) + \lambda_2 \mathcal{L}_{\texttt{perc}}(I_G, I_O),
\end{equation}
where $I_G$ and $I_O$ denote the ground truth and the generated image, respectively. We empirically set $\lambda_1 = 1.0$ and $\lambda_2 = 0.02$, and employ VGG16~\cite{VGG} as the pre-trained model for the perceptual loss.

\paragraph{Training details.} We implement our proposed method using the PyTorch framework and conduct all experiments on an NVIDIA RTX 3090 GPU. We adopt an end-to-end strategy by training both \EDN\ and \VAN~simultaneously for $150$ epochs. We adopt the Adam optimizer with an initial learning rate of $1\times10^{-4}$ and apply a cosine annealing learning rate scheduler to progressively adjust the learning rate. To improve robustness and generalization, we incorporate random cropping as the data augmentation.

\section{Experiment}\label{sec:experiment}

\begin{figure}
    \centering
    \includegraphics[width=\linewidth]{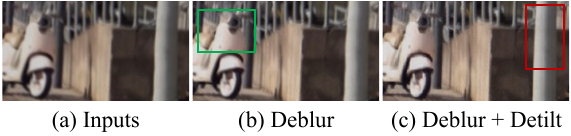}
    \caption{An example demonstrating the effectiveness of D-Net and T-Net. Given a turbulent image (a), D-Net effectively removes the blur (b). When combined with T-Net, our method corrects tilt distortions (c).}
    \label{fig:blurandtile}
\end{figure}

\begin{figure*}
    \centering
    \includegraphics[width=\linewidth]{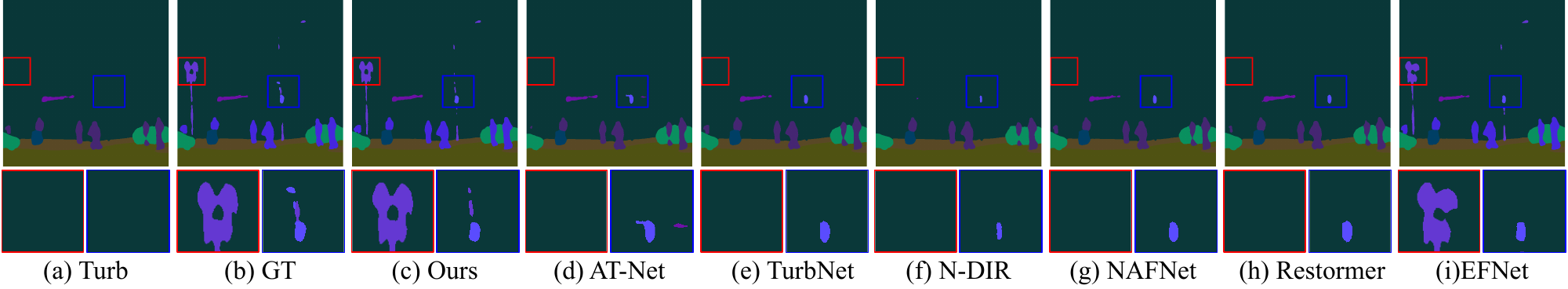}
    \caption{Visual quality comparison for semantic segmentation on the TurbEvent dataset. (a) and (b) show segmentation results from the turbulent image and the ground truth image, respectively. (c)$\sim$(i) present segmentation results from images restored using our method, AT-Net, TurbNet, N-DIR, NAFNet, Restormer, and EFNet. Close-up views are provided below each image. Additional results are provided in the supplementary material.}
    \label{fig:seg}
\end{figure*}

\subsection{Evaluation on TurbEvent dataset}
For the training and testing data split in the TurbEvent dataset, we first utilize image labels to categorize the data into multiple groups. Within each group, we randomly split the data into training and testing subsets with a $9:1$ ratio. This strategy ensures a similar distribution between the training and testing datasets. As a result, we obtain $5665$ pairs for training and $653$ pairs for testing.

We compare our method against three image-based turbulence removal approaches, two general image restoration methods, and one event-based deblurring method: AT-Net~\cite{AT-Net}, TurbNet~\cite{TurbNet}, N-DIR~\cite{Li_2021_NIDR}, NAFNet~\cite{NAFNet}, Restormer~\cite{Zamir2021Restormer}, and EFNet~\cite{sun2022efnet}. Additionally, for the general image restoration method Restormer, we also feed event data alongside image frames into the network, denoted as Restormer$^*$. All supervised methods~\cite{AT-Net, TurbNet, NAFNet, Zamir2021Restormer, sun2022efnet} are re-trained on our TurbEvent dataset for fair comparison. The quality of restored images is evaluated using three commonly adopted metrics: PSNR, SSIM, and LPIPS.

Quantitative results are presented in \Cref{tab:heat}, while qualitative comparisons are shown in \Cref{fig:realdata}. As demonstrated, our method consistently achieves the highest performance across all three metrics. In terms of PSNR, event-guided methods notably surpass all other image-based methods.

For visual quality, our method effectively reduces image blur compared to single-image based techniques. Furthermore, our approach reconstructs images with fewer artifacts and sharper details than event-based deblurring method EFNet. For Restormer$^*$, the straightforward concatenation strategy results in only a slight improvement in performance. Notably, as shown in the first row of  \Cref{fig:realdata}, only our method effectively addresses pixel displacement.

\begin{table}
		\centering
		\small
        \caption{Quantitative comparison of the semantic segmentation task on restored images from our TurbEvent dataset.}
		\label{tab:seg}
		\begin{tabular}{lcccc}
			\toprule
            & ~Event~ & ~mIoU $\uparrow$~  & ~Dice $\uparrow$~ & ~FWIoU $\uparrow$~\\
			\midrule
            Turb &  & 0.923 & 0.959  &  0.931 \\
            Ground truth &  & 0.941 & 0.969  & 0.948  \\
            \midrule
			AT-Net & \xmark & 0.922 & 0.958  & 0.932  \\
			TurbNet & \xmark & 0.926 & 0.961 & 0.936 \\
			N-DIR & \xmark  & 0.910  & 0.951 &  0.921 \\
            NAFNet  & \xmark &  0.923 & 0.959 & 0.936  \\
            Restormer & \xmark & 0.928 & 0.961 & 0.937  \\
            EFNet & \cmark  & 0.930 & 0.963 & 0.939 \\
			Ours & \cmark & 0.936 & 0.966 & 0.943  \\
			\bottomrule
		\end{tabular}
\end{table}

\subsection{Evaluation on real-captured data}
We captured real-world turbulent images using a hybrid camera system\footnote{Detailed configuration can be found in the supplementary.}, with results shown in \Cref{fig:realdata2}. As illustrated, our method consistently outperforms other state-of-the-art approaches on real-world data, restoring more details and demonstrating superior generalization capabilities.

To demonstrate the effectiveness of D-Net and T-Net, designed for deblurring and detilting respectively, we present results obtained using D-Net alone. As shown in \Cref{fig:blurandtile}, D-Net effectively removes blur (green box). When combined with T-Net, our method addresses tilt issues (red box). D-Net leverages residual learning with global connections, a technique well-established for image deblurring tasks. Conversely, T-Net estimates a tilt flow field to accurately model geometric pixel shifts.

\subsection{Evaluation on downstream task}
To comprehensively assess the quality of restored images, we evaluate the performance improvement that EvTurb brings to image-based semantic segmentation. The original ADE20K dataset~\cite{zhou2019semantic, zhou2017scene} provides ground truth labels for semantic segmentation. We select InternImage-XL~\cite{Segmentation} as the benchmark model, which leverages deformable convolutions to enable adaptive spatial aggregation and effective receptive fields.

Three common metrics are used for evaluation: mean Intersection over Union (mIoU), Dice coefficient, and frequency-weighted Intersection over Union (FWIoU). The quantitative results, presented in \Cref{tab:seg}, demonstrate that our method achieves the highest improvements compared to other methods, approaching the performance on ground-truth images. The visual comparison in \Cref{fig:seg} shows that our method enables more accurate image segmentation.\footnote{Note that our train and test split differs from original ADE20K, so absolute values are not directly comparable to those of the original dataset. But the difference between restored and turbulent images demonstrate our method enhances downstream performance.}

\subsection{Ablation studies}
We conducted four ablation studies to evaluate the contributions of each module in our EvTurb method, as summarized in \Cref{tab:ablation}. Specifically, we first removed the EGDC blocks (denoted as ``W/o EGDC'') to investigate their role in data fusion between two modals. Next, we removed the \VAN~to assess its effectiveness in mitigating tilt distortion (denoted as ``W/o \VAN''). Then, we excluded the \VAN~to directly evaluate its specific impact on tilt correction (denoted as ``W/o Variance''). Lastly, we remove the flow-based architecture for directly restoring images to determine its role in refinement (denoted as ``W/o Flow''). Results demonstrate that our complete model achieves better overall performance. The model struggles with multi-scale image event data integration without EGDC blocks. Without the \VAN~module, results exhibit significant tilt degradation.

\begin{table}
		\centering
		\small
        \caption{Quantitative results of ablation study.}
		\label{tab:ablation}
		\begin{tabular}{lccc}
			\toprule
            & ~~~PSNR $\uparrow$~~~  & ~~~SSIM $\uparrow$~~~ & ~~~LPIPS $\downarrow$~~~ \\
			\midrule
            W/o EGDC & 28.08 & 0.7957 & 0.2925 \\
            W/o Variance & 28.42 & 0.8129 & 0.2384 \\
            W/o T-Net  &  27.70 & 0.7524 &  0.3352 \\
            W/o Flow  & 28.16  & 0.8088 & 0.2553 \\
			Our full model & 29.67 & 0.8405 & 0.2010\\
			\bottomrule
		\end{tabular}
\end{table}
\section{Conclusion}\label{sec:conclusion}
In this paper, we propose EvTurb, a novel framework for atmospheric turbulence removal that leverages high-speed event streams. By modeling the blur operator based on event-recorded intensity changes and the tilt operator through event-triggering frequency, we effectively decouple blur and tilt distortions. We further perform turbulence removal through a structured two-step framework. Evaluations on our real-captured TurbEvent dataset demonstrate that our method significantly outperforms other methods while maintaining competitive runtime efficiency.

\paragraph{Limitations.} A limitation of our approach stems from potential misalignment between the event camera and the frame camera. Although we employ precise calibration and a beam-splitting optical setup to alleviate this issue, minor misalignment may still persist, potentially affecting the accuracy of data fusion. Additionally, our implementation assumes identical resolutions for both cameras, simplifying processing but overlooking cross-resolution discrepancies commonly encountered in real-world applications. Furthermore, the monochromatic nature of event data introduces a domain gap when fused with RGB images.

{
    \small
    \bibliographystyle{ieeenat_fullname}
    \bibliography{main}
}

\end{document}